\documentclass{article}

\usepackage{spconf,amsmath,graphicx}
\usepackage{url}            

\title{Deep Learning for Lagrangian drift simulation at the sea surface}
\name{Daria Botvynko$^{1}$, Carlos Granero-Belinchon$^{2}$, Simon van Gennip$^{3}$, Abdesslam Benzinou$^{1}$, Ronan Fablet$^{2}$\thanks {This work was supported by LEFE program (LEFE MANU and IMAGO projects IA-OAC), CNES (OSTST DUACS-HR and SWOT ST DIEGO) and ANR Projects Melody (ANR-19-CE46-0011) and OceaniX (ANR-19-CHIA-0016). It benefited from HPC and GPU resources from Azure (Microsoft Azure grant) and from GENCI-IDRIS (Grant 2021-101030)}}
\address{$^1$ ENIB, Lab-STICC, UMR CNRS 6285, 29238 Brest \\ $^2$ IMT Atlantique, Lab-STICC, UMR CNRS 6285, 29238 Brest, France \\ $^3$ Mercator Ocean International, 31400 Toulouse, France}

\begin{document}

\ninept

\maketitle

\begin{abstract}
We address Lagrangian drift simulation in geophysical dynamics and explore deep learning approaches to overcome known limitations of state-of-the-art model-based and Markovian approaches in terms of computational complexity and error propagation. We introduce a novel architecture, referred to as DriftNet, inspired from the Eulerian Fokker-Planck representation of Lagrangian dynamics. Numerical experiments for Lagrangian drift simulation at the sea surface demonstrates the relevance of DriftNet w.r.t. state-of-the-art schemes. Benefiting from the fully-convolutional nature of DriftNet, we explore through a neural inversion how to diagnose model-derived velocities w.r.t. real drifter trajectories. 
\end{abstract}


\section{Introduction}

\label{sec:intro}

The modeling and simulation of Lagrangian drift at sea surface is a key scientific challenge in operational oceanography \cite{op_oceanography_currents}. Applications of interest range from  plastic and other debris tracking \cite{marine_debris_lagr}, algae or plankton drift \cite{algae_lagr}, to the forecasting of drifting objects for search and rescue operations \cite{search_and_rescue}. The analysis of Lagrangian drifts is also widely used to diagnose ocean numerical models w.r.t. Lagrangian trajectories observed in the field \cite{ogcm_evaluation_drifters}.

From a methodological point of view, we may distinguish two broad categories of approaches for Lagrangian drift simulation: model-based approaches~\cite{lagr_model_plastic} and Markov data-driven schemes~\cite{marine_litter_lagr}. Model-based approaches rely on the implementation of a sequential advection process knowing the underlying velocity fields~\cite{model_lagr_adv_veloc}. Due to its sequential nature, these schemes are not highly-scalable to simulate large ensembles of drift trajectories. Besides, small errors in the underlying velocity fields may strongly affect the trajectory patterns. The latter is a critical issue as, for operational applications, these velocity fields are prone to inversion errors. By contrast, Markov data-driven schemes naturally account for uncertainties in the drift process but mainly apply to relatively coarse space-time resolutions. Their extension to fine-scale patterns is a challenge, \cite{markov_appli}. We also note that in both approaches the simulation relies at each time-step on a location-wise velocities only, which may only apply for smooth velocity fields.

Recently, a growing literature has emerged for applications of deep learning to trajectory data, including among others, generative models \cite{gan_amedee}, \cite{traj_predict}, \cite{gan_lagrangian}, \cite{cond_gan_traj}, \cite{cond_gen_traj}, short-term forecasting \cite{traj_pred_rnn} and classification issues \cite{classif_dl_images}. Most of the studies using generative approaches rely on sequential approaches and recurrent neural networks, which may face the same limitations as the ones identified above for the state-of-the-art schemes for Lagrangian drift simulation.
To our knowledge, only very few previous studies have explored the conditional  simulation of Lagrangian drift \cite{cond_eulerian_lagr}.
We may however note that some studies stress the potential for considering higher-dimensional latent representation of trajectory data \cite{duong_geotracknet}. 

Here, we propose a novel deep learning scheme for the conditional simulation of Lagrangian drifts. We introduce a convolutional architecture with a spatially-explicit latent representation, inspired by Eulerian Fokker-Plank representations of drift processes. It results in a non-local simulation of drift trajectories conditionally to velocity fields. 
In our numerical experiments, we consider an application to Lagrangian drift simulation at the sea surface for a case-study region off the coast of California. The reported results support the relevance of this approach. Through an inverse problem formulation, we further illustrate how our approach provides new means to inform sea surface velocities from real drift trajectories.

The remainder is organized as follows. Section \ref{sec:problem_statement} provides some background. We introduce the proposed deep learning approach in Section \ref{sec:proposed_method} and report numerical experiments in Section \ref{sec:results}. Section \ref{sec:conclusion} discusses further our main contributions.

\section{Problem statement}

\label{sec:problem_statement}

Lagrangian fluid dynamics consider the fluid as an ensemble of particles and describe the flow properties trough the trajectories of the particles advected by the flow \cite{lagrangian_flow}. This results in  particle trajectories governed by the following ordinary differential equation: 
\begin{equation}\label{eq:advection}
    \frac{\partial{\vec{\small{r}}}(\vec{\small{r_0}},t)}{\partial t} = \vec{\small{v}}(\vec{\small{r_0}},t)= \vec{\small{u}}(\vec{\small{r}}(t),t)
\end{equation}
where ${\vec{\small{r}}}(\vec{r_{0}}, t)$ and $\vec{\small{v}}(\vec{r_{0}},t)$ are respectively the position and velocity at time $t$ of the particle initially located at $\vec{\small{r_0}}$, and $\vec{\small{u}}(\vec{\small{r}}(t),t)$ is the Eulerian velocity of the flow at position $\vec{\small{r}}(t)$ and time $t$. State-of-the-art model-driven schemes implement such Lagrangian drift dynamics using explicit time integration schemes, such as Runge-Kutta 4. The sequential nature of the above equation clearly propagates and accumlulates errors if the underlying velocities are not perfectly known \cite{seeding_radius_lagr_drift_error}. Besides, from a computational point of view, the location-dependent parameterization of the right-hand term in eq.(\ref{eq:advection}) leads to a relatively poor parallelization potential for the simulation of large ensembles of trajectories.

Given eq.(\ref{eq:advection}), learning-based approaches for Lagrangian drift simulation naturally consider recurrent neural networks as investigated in \cite{traj_drifters_rnn} using LSTM and broadly used from trajectory simulations \cite{traj_pred_rnn}.
This is also the theoretical background for Markov models \cite{markov_hidden} which can regarded as an extension of eq.(\ref{eq:advection}) accounting for uncertainties. 
Both categories of approaches state the targeted issue as the simulation of a time sequence of positions $\vec{\small{r}}(\vec{\small{r_0}}, t)$ according to a discrete-time generalization of eq.(\ref{eq:advection}):
\begin{equation}\label{eq:markovian dyn}
    \vec{\small{r}}(t+\delta) = \mathcal{F} \left( \vec{\small{r}}(t) , \mathbf{z}_t,t\right )
\end{equation}
with $\delta$ the simulation time step,
$\mathcal{F}$ some deterministic or stochastic operator to be trained and $\mathbf{z}_t$ conditioning variables. In general $\mathbf{z}_t=\vec{\small{u}}(\vec{\small{r}}(t),t)$. This representation suffers from the same limitation as 
eq.(\ref{eq:advection}) in terms of computational complexity and error propagation.

Interestingly, we can also derive an Eulerian formulation of Lagrangian dynamics. Through Fokker-Planck formalism \cite{lagrangian_fokker_planck_plankton}, we can 
generalize eq.(\ref{eq:advection}) to the time propagation of the pdf of the moving particle as follows:
\begin{equation}\label{eq:fokker-planck}
    \frac{\partial}{\partial t} p_{\vec{\small{r}}}(x, t) = - \frac{\partial}{\partial x}[\mu(x, t)p_{\vec{\small{r}}}(x, t)]
\end{equation}
where $\small{x}$ represents spatial position in the Eulerian framework, $t$ the time, $p_{\vec{\small{r}}}(x, t)$ the probability density function of the position $\vec{\small{r}}$ of the moving particle and $\mu(x, t)$ a drift field. In our case, this drift term relates to the underlying velocity field $\vec{\small{u}}(x,t)$. Solving eq.(\ref{eq:fokker-planck}) relies on a time integration scheme from some initial pdf $p_{\vec{\small{r}}}(x, t = 0)= p_{\vec{\small{r_0}}}(x,0)$. Computationally speaking, this Eulerian representation is no more state-dependent as in eq.(\ref{eq:advection}) but a partial differential equation (PDE). This is particularly appealing from a learning point of view as numerous studies have stressed the relevance of convolutional architectures as PDE solvers \cite{PDE_Net}. 

Here, we draw on this Eulerian Fokker-Planck representation of Lagrangian dynamics to explore Eulerian convolutional  neural architectures for Lagrangian drift simulation.

\section{Proposed method}
\label{sec:proposed_method}

This section describes the proposed deep learning approach for Lagrangian drift simulation, referred to as DriftNet. We first introduce our neural architecture before the learning setup and the neural inversion approach. 

\subsection{DriftNet architecture}
Formally, let us introduce $\mathcal{D}$ the spatial domain of interest and $\mathbf{u}=\{\mathbf{u}_{t_0},\mathbf{u}_{t_0+\Delta},\ldots,\mathbf{u}_{t_0+K\Delta}\}$ a sequence of velocity fields over domain $\mathcal{D}$ from time $t_0$ to $t_0+K\Delta$ with time resolution $\Delta$ and number of time steps $K$. Similarly, we denote by $\vec{r}=\{\vec{r}_{t_0},\vec{r}_{t_0+\Delta},\ldots,\vec{r}_{t_0+K\Delta}\}$ a Lagrangian drift given as a time series of locations in $\mathcal{D}$ from $t_0$ to  $t_0+K\Delta$.   

Inspired by Fokker-Plank representation eq.(\ref{eq:fokker-planck}), we consider the following latent representation for a Lagrangian drift $\vec{r}$:
\begin{equation}
\left \{
    \begin{array}{ccl}
         \mathbf{y} &=&  \mathcal{E} \left ( \mathbf{u}, \mathbf{y}_0 \right ) \\
         \vec{r}&=&  \mathcal{M} \left (\mathbf{y} \right ) \\
    \end{array}\right.
\end{equation}
where $\mathbf{y}$ is a space-time-explicit latent embedding of $\vec{r}$, and $\mathbf{y}_0$ some initial latent embedding to encode the known initial position $\vec{r}_{t_0}$. 
Eulerian neural operator $\mathcal{E}$ computes latent embedding $\mathbf{y}$ given velocity conditions $\mathbf{u}$ and initial representation  $\mathbf{y}_0$. Neural operator $\mathcal{M}$ maps 
latent representation $\mathbf{y}$ to the targeted Lagrangian drift $\vec{r}$. As such, it aims at mapping a multi-dimensional tensor to a time series of locations in 
$\mathcal{D}$. 

Regarding the parameterization of operator $\mathcal{E}$, we do not constrain the latent representation to be univariate similarly to a pdf in Fokker-Planck formulation. We explore the computational efficiency of deep learning schemes for higher-dimensional latent representations. Overall, operator $\mathcal{E}$ is composed of four elementary units: a series of 3{\sc d} convolution layers (with the number of channels evolving as follows: $2 \rightarrow 11 \rightarrow 16$) with a non-linear LeakyReLU activation, a 2{\sc d} convolutional LSTM block and the last series of 3{\sc d} convolution layers (number of channels evolving as follows: $16 \rightarrow 8$) with a non-linear LeakyReLU activation. Mapping operator $\mathcal{M}$ then combines a {\em Softmax} layer and a last block of 1{\sc d} convolutional layers (number of channels evolving as follows: $8 \rightarrow 2$) to transform latent variable $\mathbf{y}$ into a set of positions $\vec{r}$.

The initial latent representation  $\mathbf{y}_0$ provided to operator $\mathcal{E}$ encodes the initial position of the particle. For time $t_0$, we consider 
a 2{\sc d} map where the value at each point of the reference domain $\mathcal{D}$ is given by a normalized version of the distance to the initial position of the particle  $\vec{r}_{t_0}$. For the next time steps the encoded normalized distances from the initial position to each grid of the map are attenuated by the dispersion factor computed for the area of study (with regards to the maximum speed in the zone). 

Through the different convolutional blocks and the convolutional LSTM unit, we expect the resulting architecture to capture relevant information at different space-time scales, and not only in a point-wise manner, as would be the case for the straightforward implementation of Fokker-Plank representation eq.(\ref{eq:fokker-planck}). 

\subsection{Learning setting}
\label{sec:proposed_methods:learning_setting}

We consider a supervised training of the proposed neural architecture according to the following two losses: 
\begin{enumerate}
    \item the Mean Square Error (MSE) between the reference and simulated trajectories for the considered $K$-step simulation of $N_{T}$ particles
\begin{equation}\label{eq:lossmse}
    \mathcal{L}_{MSE}=\frac{1}{N_{T}K} \sum_{i=0}^{N_{T}} 
    \sum_{t = 0}^{K} (\vec{r}_{R}(\vec{r}_{0,i},t) - \vec{r}_{S}(\vec{r}_{0,i},t))^2
\end{equation}
   
    \item Liu index between the reference and simulated trajectories \cite{liu2011evaluation}
\begin{equation}\label{eq:lossliu}
    \mathcal{L}_{Liu}=\frac{1}{N_{T}} \sum_{i=1}^{N_{total}} \frac{\sum_{j=1}^{K} d_{i_{j}}}{\sum_{j=1}^{K} l_{i_{j}}}
\end{equation}
\noindent where $d_{i_{j}}$ is the Euclidean distance between the reference and the simulated trajectories number $i$ at time step $j$ and $l_{i_{j}}$ is the length of reference trajectory $i$ between the initial position and the position at time step $j$.
\end{enumerate}
Overall, the training loss $\mathcal{L}$ is a weighted sum of these losses: $\mathcal{L}=\alpha \cdot \mathcal{L}_{MSE} + \beta \cdot \mathcal{L}_{Liu}$.
From cross-validaton experiments, we set $\alpha$ and $\beta$ to
0.2 and 0.8. Using Pytorch\footnote{The open source code ofis available at \url{https://github.com/CIA-Oceanix/DriftNet}}, our learning setup relies on Adam optimizer with a learning rate of 5e-3 over 100 epochs. 

\subsection{Neural inversion of sea surface currents}
\label{sec:problem_statement:inversion}
Contrary to previous works, the proposed neural architecture for Lagrangian drift simulation does not rely on a point-wise relationship but may exploit space-time features of interest in the conditioning velocity fields. Here, we explore how a trained DriftNet  can provide new means to diagnose model-driven velocities w.r.t. real Lagrangian trajectories. 

Let us consider a trained DriftNet with operators $\{\mathcal{E}^*,\mathcal{M}^*\}$ and a real Lagrangian trajectory $\vec{r}^{R}$. We aim at identifying the correction $\widehat{d}{\mathbf{u}}$ to the conditioning velocity fields ${\mathbf{u}}$ such that DrifNet trajectory simulated with velocity condition ${\mathbf{u}} + \widehat{d}{
\mathbf{u}}$ 
and initial poition $\vec{r}^{R}_{t_0}$ best matches trajectory $\vec{r}^{R}$. This leads to the following minimization problem:
\begin{equation}
\label{eq: inversion}
\widehat{d}{\mathbf{u}} = \arg \min_ {d\mathbf{u}}
    \lvert\lvert {\vec{r}}^{R} - \mathcal{M}^{*}(\mathcal{E}^{*}(\mathbf{u} + d\mathbf{u}, \mathbf{y}_0)) \rvert\rvert ^ 2
\end{equation}
Thanks to the automatic differentiation natively embedded in deep learning framework, we directly solve this minimization using a fixed-step gradient descent, typically over 200 gradient steps with stepsize of 5e-2.

\section{Results and Discussion}
\label{sec:results}
In this section we introduce our operational oceanography case study. We present DriftNet results for Lagrangian drift simulation at sea surface  and compare them to those of state-of-the-art models. We also report neural inversion examples for the retrieval of corrected sea surface velocities.

\subsection{Case study and Datasets}
\label{sec:results:case_study}

\begin{figure}[h]
    \includegraphics[width = 0.45\textwidth]{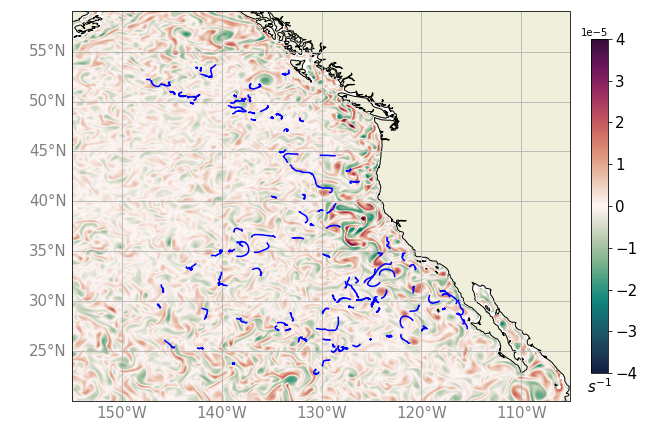}
    \caption{ Snapshot of 9-days GDP drifters trajectories deployed close in time (blue) superimposed on relative vorticity of Glory12v1 velocity field for the case-study area on 20/10/1993.}
    \label{fig:curl_glo12_15m_nep}
\end{figure}

The focus region for our Lagrangian drift simulations is the California Current System (CCS) within the North East Pacific (20°N to 60°N and 160°E to 110°E, see Fig.\ref{fig:curl_glo12_15m_nep}). 
This zone is a major eastern boundary upwelling system wherein cold nutrient-rich water upwelled along the coast fuel one of the most productive marine ecosystem. At the surface it is crossed by the equatorward-flowing California Current (CC) and is a part of the anticyclonic North Pacific gyre  \cite{california_current}. It is characterised by intense mesoscale actvity with long-lived eddies generated along the coast travelling offshore westward. The strong mesoscale and submesoscale dynamics of the region together with the presence of a large number of drifters \cite{elipot_drifters} make it particularly well suited for our study.

Two datasets are used:
Global Ocean General Circulation Model reanalysis product GLORYS12V1 and
Drifters trajectories records from the Copernicus Marine Services dataset.
GLORYS12V1 \footnote{https://doi.org/10.48670/moi-00021} is a reanalysis product of the E.U. Copernicus Marine Service with eddy-resolving resolution of 1/12° and 50 vertical layers \cite{lellouche_glorys12} from 1993 to present. We focus here on GLORYS12V1 sea surface velocities.

Satellite-tracked surface drifting buoys, referred to as drifters, data is collected from the Surface Drifter Data Assembly Centre (NOAA AOML), participating in Global Drifter Program (GDP) and is distributed by Copernicus E.U. Marine Service \footnote{https://doi.org/10.17882/86236}. Drifters trajectories records provide sea surface currents observations gridded at a 6-hour resolution, from 1990 to present for the entire globe \cite{elipot_drifters}, \cite{drifters_lumpkin}. The mean trajectory length of drifters deployed in the zone of study is 280 days. As the focus here is set on short-time series, trajectories have been divided into 9-day segments. Overall, our dataset comprises a total of 12.436 9-day trajectories segments with a 6-hour resolution from 1993 until 2020.

We generate a synthetic dataset of Lagrangian drift trajectories according to GLORYS12V1 sea surface velocity fields using as initial positions those of the 9-day trajectories of real drifters. This dataset generated with Ocean Parcels tool \cite{model_lagr_adv_veloc} leads to a dataset of 12.436 9-day simulated trajectories with a 6-hour resolution. We use this dataset to asses the proposed deep learning scheme according to the following train-validation-test split: we randomly select 80\% of the trajectories as training data, and the remaining 20\% are equally split into the validation and test datatests.

\subsection{Benchmarking experiments}
\label{sec:results:benchmark}

Our numerical experiments benchmark DriftNet (amounting to 2.230.491 trainable parameters) with the following state-of-the-art neural architectures:
\begin{itemize}
    \item a LSTM architecture with a one-dimensional latent space, inspired from \cite{traj_pred_rnn} ( 1.185.448 trainable parameters);
    \item a 2{\sc d} Convolutional LSTM architecture with an additional  ConvTranspose layer to map the 2{\sc d} LSTM latent space to a sequence of positions (980.410 trainable parameters);
    \item a CNN architecture inspired from \cite{traj_predict}, which combines 3{\sc d} convolutional blocks and ConvTRanspose layers  to map the embedded vector to the targeted temporal resolution (2.069.718 trainable parameters).
\end{itemize}

As evaluation metrics, we consider the Liu index introduced in eq.(\ref{eq:lossliu}), and the RMSE (root mean square error), defined as the root of the MSE defined in eq.(\ref{eq:lossmse}), between the reference and the simulated positions ($\gamma(\vec{r}_{t})$) and Lagrangian time scales ($\gamma(\mathcal{T}$)) \cite{lagr_time_scale}. 
The Lagrangian time scale, $\mathcal{T}=\int R_{\vec{v}}(\tau) d\tau$, with $R_{\vec{v}}(\tau)$ the autocorrelation function, is a measure of the time during which the velocities are correlated with themselves .

\begin{figure}[h]
    \centering
    \includegraphics[width = \linewidth]{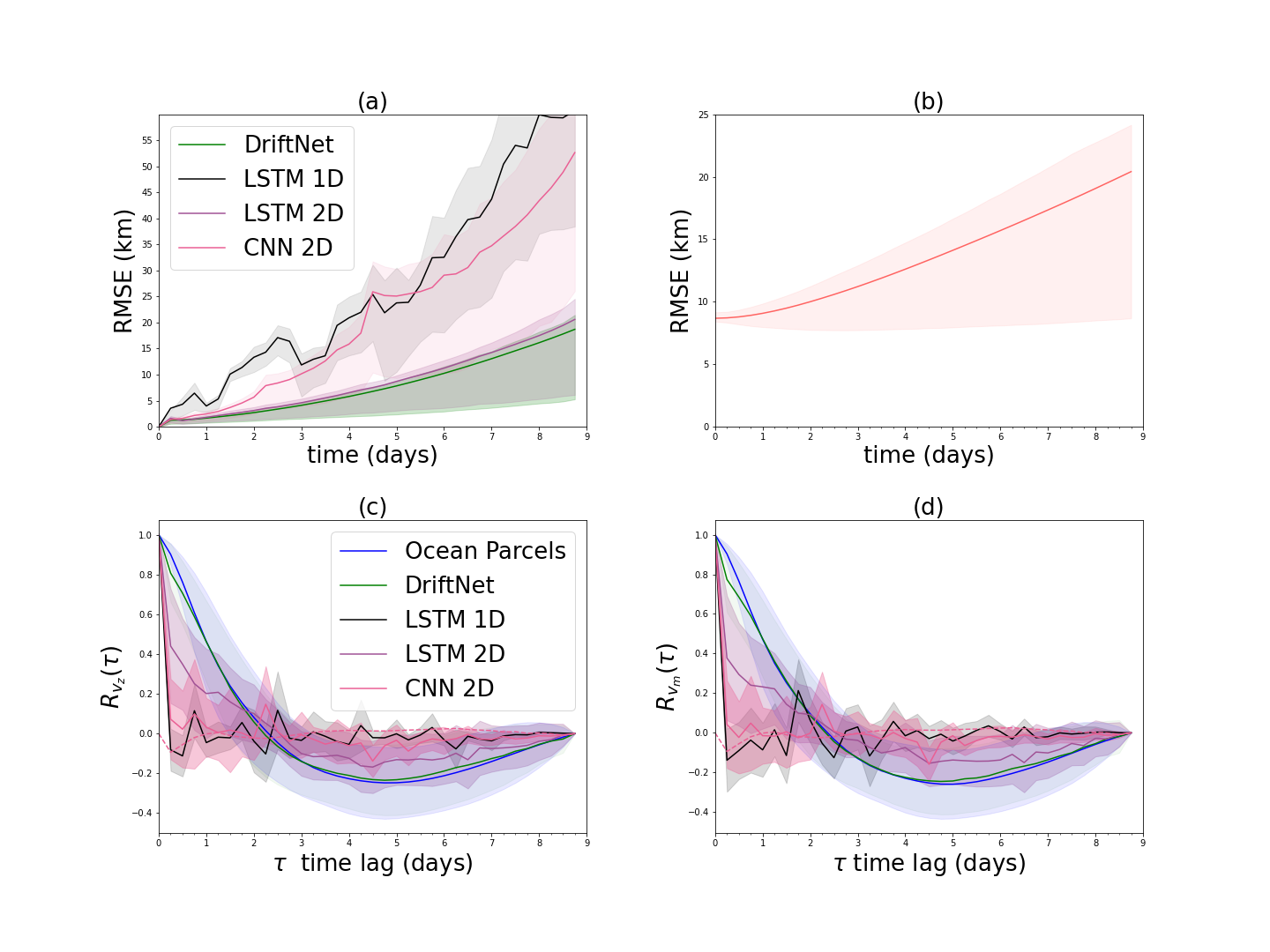}
    \caption{(a) Mean separation distance in km between trajectories simulated with Ocean Parcels, DriftNet, LSTM and CNN models as a function of time in days. (b) Distance in km between trajectories seeded in drifters launching positions and trajectories seeded in the 1/12° radius around them, both simulated with Ocean Parcels.  The shaded area correspond to the distances comprised between the first and the third quantile. (c), (d) Autocorrelation for zonal (left) and meridional (right) velocities as a function of time lag ($\tau$) for different simulation methods and Ocean Parcels as reference.}
    \label{fig:dist_and_R}
\end{figure}

Figure~\ref{fig:dist_and_R} shows separation distance statistics between simulated and reference Lagrangian drifts for the benchmarked models. DriftNet clearly outperforms the other deep learning models. It reduces mean separation distance at 9-th day by 66\% compared to LSTM-1D, 5\% compared to LSTM-2D model and by 65\% compared to CNN-2D (see Table \ref{table:metrics}). Importantly, the simulation performance is of the same order as the one corresponding to Ocean Parcels with a initial position uncertainty of 1/12$^\circ$ (Fig.\ref{fig:dist_and_R}b). This is a direct consequence of the chaos in turbulent flows, which is widely documented \cite{seeding_radius_lagr_drift_error}. Similar conclusions can be drawn for Liu index, which provides an assessment of  
trajectory geometry similarities. DrifNet outperforms the second best model by 23\%. 

We analyse further the relevance of DriftNet simulations in terms of auto-correlation patterns (Fig.\ref{fig:dist_and_R}c-d). We report a very good match between DriftNet simulations and reference Lagrangian drifts. These results are in line withe the good performance in terms Langrangian time scales discussed above.

\begin{table}[h]
 
    \centering
        \begin{tabular}{|c|c|c|c|c|}
            \hline
            Model &  $\gamma$($\mathcal{T}$), days & $\text{Liu}$ & $\gamma(\vec{r}_{t = 9})$, km \\
            \hline
            LSTM 1D &  3.74 & 0.8 & 59.2 \\
            LSTM 2D &  1.78 & 0.26 & 20.7\\
            CNN 2D &   12 & 0.64  & 52\\
            DriftNet & 0.7 & 0.2 & 19.6 \\
            \hline
        \end{tabular}
    \centering
    \caption{Metrics comparison between DriftNet and state-of-the-art methods. The used metrics are the Liu index, the RMSE between positions $\gamma(\vec{r}_{t = 9})$, and Lagrangian time scales $\gamma(\mathcal{T}$). Each metric is averaged over the trjaectories of the test dataset.}
    \label{table:metrics}
\end{table}

\subsection{Inversion examples}
\label{sec:results:inversion} 

\begin{figure}[h]
    \centering
    \includegraphics[width = \linewidth]{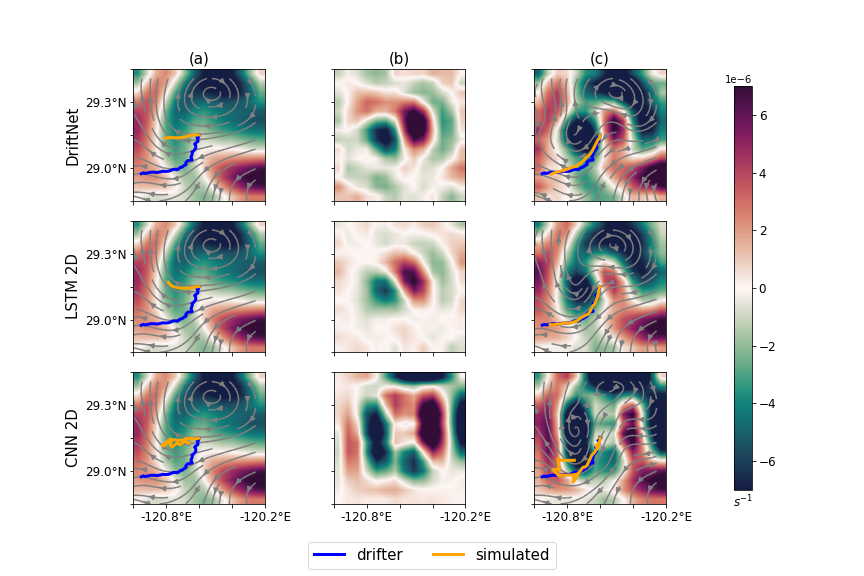}
    \caption{Neural inversion examples for velocity anomaly fields from real drifter trajectories: vorticity of the  velocity field  at time $t_0$ with the reference (blue) and simulated (orange) drift (a), vorticity of the estimated velocity anomaly field at time $t_0$ (b) and vortcity of the corrected velocity field  at time $t_0$ with reference (blue) and simulated (orange) drift. The first three rows relate to the first example using different deep learning architectures in inversion scheme eq.(\ref{eq: inversion}), and the last three to the second example.}
    \label{fig:backprop}
\end{figure}

We report inversion examples using the approach described in Section \ref{sec:problem_statement:inversion}. As illustrated in Fig.\ref{fig:backprop}, we observe a clear mismatch between the real drifter trajectory and the one simulated from GLORYS12V1 sea surface velocities. It reveals that GLORYS12V1 product cannot recover all fine-scale patterns from the assimilation of available satellite-derived observation datasets \cite{lellouche_glorys12}. 

As reported for the two examples in Fig.\ref{fig:backprop}, DrifNet-based neural inversion identifies velocity anomalies such that the simulated trajectories match closely the real ones. By contrast, this inversion scheme does not apply relevantly when considering the other neural architectures. 
DrifNet-based velocity anomalies exhibit fine-scale vorticity patterns. Such structures are too small to be well resolved by reanalysis products which do not capture the small meso and submesoscale \cite{lellouche_glorys12}. Equally, limitations also exist in resolving the upper mesoscale with for instance eddy features potentially poorly positioned, mis-dimensioned, or incoherent over time, all of which having a direct impact on the Lagrangian transport \cite{ogcm_evaluation_drifters}. 

\begin{figure}[h]
    \centering
    \includegraphics[width = \linewidth]{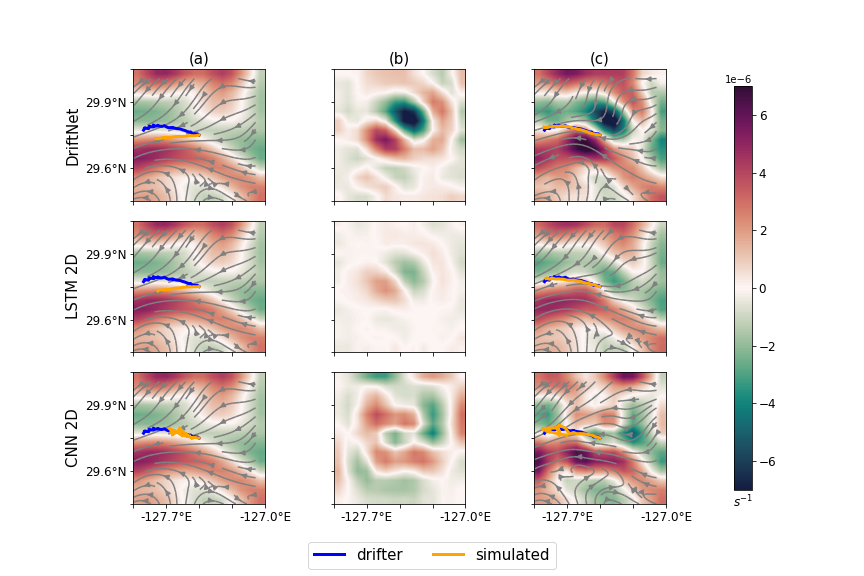}
\end{figure}

\section{Conclusions}
\label{sec:conclusion}
In this study we proposed a novel method for Lagrangian drift simulation and analysis at sea surface based on a Deep Learning scheme. It relates to  Eulerian Fokker-Planck interpretation of Lagrangian dynamics. We introduce a fully-convolutional architecture and an Eulerian latent representation of Lagrangian trajectories.
The resulting Lagrangian drift simulations outperform previously explored state-of-the-art methods for trajectory data. We also point out its potential to diagnose sea surface velocities derived by operational models w.r.t. real drifter trajectories.

We believe this study to open new research avenues for the development of deep learning schemes for Lagrangian drift simulation. From a methodological point, the extension of the proposed architecture to a conditional GAN setting \cite{cond_gan_mirza} naturally arises to account the variability of real drift trajectories. Regarding operational oceanography, we may envision extensions of the proposed scheme with satellite-derived observations and other model-driven variables, including among others sea surface temperature fields, satellite altimetry data, wave models' outputs... 
Our neural scheme may also provide new means in movement ecology to study the interactions between animal movement trajectories and their environment \cite{animal_traj_deep_learning}. 


\label{sec:refs}

\bibliographystyle{IEEEbib}
\bibliography{biblio}

\begin{thebibliography}{10}

\bibitem{op_oceanography_currents}
J.~R{\"o}hrs, G.~Sutherland, G.~Jeans, M.~Bedington, A.~K. Sperrevik, K.-F.
  Dagestad, Y.~Gusdal, C.~Mauritzen, A.~Dale, and J.~H LaCasce,
\newblock ``Surface currents in operational oceanography: Key applications,
  mechanisms, and methods,''
\newblock {\em Journal of Operational Oceanography}, 2021.

\bibitem{marine_debris_lagr}
N.~Maximenko, J.~Hafner, and P.~Niiler,
\newblock ``Pathways of marine debris derived from trajectories of lagrangian
  drifters,''
\newblock {\em Marine pollution bulletin}, vol. 65, 2012.

\bibitem{algae_lagr}
Y.~B. Son, B-J. Choi, Y.~H. Kim, and Y.-G. Park,
\newblock ``Tracing floating green algae blooms in the yellow sea and the east
  china sea using goci satellite data and lagrangian transport simulations,''
\newblock {\em Remote Sensing of Environment}, vol. 156, 2015.

\bibitem{search_and_rescue}
{\O}.~Breivik, A.~A. Allen, C.~Maisondieu, and M.~Olagnon,
\newblock ``Advances in search and rescue at sea,''
\newblock {\em Ocean Dynamics}, vol. 63, pp. 83--88, 2013.

\bibitem{ogcm_evaluation_drifters}
C.~N Barron, L.~F Smedstad, J.~M Dastugue, and O.~M. Smedstad,
\newblock ``Evaluation of ocean models using observed and simulated drifter
  trajectories: Impact of sea surface height on synthetic profiles for data
  assimilation,''
\newblock {\em Journal of Geophysical Research: Oceans}, vol. 112, 2007.

\bibitem{lagr_model_plastic}
S.~Liubartseva, G.~Coppini, R.~Lecci, and E.~Clementi,
\newblock ``Tracking plastics in the mediterranean: 2d lagrangian model,''
\newblock {\em Marine pollution bulletin}, vol. 129, 2018.

\bibitem{marine_litter_lagr}
E.~Zambianchi, M.~Trani, and P.~Falco,
\newblock ``Lagrangian transport of marine litter in the mediterranean sea,''
\newblock {\em Frontiers in Environmental Science}, vol. 5, 2017.

\bibitem{model_lagr_adv_veloc}
M.~Lange and E.~van Sebille,
\newblock ``Parcels v0. 9: prototyping a lagrangian ocean analysis framework
  for the petascale age,''
\newblock {\em Geoscientific Model Development}, vol. 10, 2017.

\bibitem{markov_appli}
S.~Fine, Y.~Singer, and N.~Tishby,
\newblock ``The hierarchical hidden markov model: Analysis and applications,''
\newblock {\em Machine learning}, vol. 32, 1998.

\bibitem{gan_amedee}
A.~Roy, R.~Fablet, and S.~L. Bertrand,
\newblock ``Using generative adversarial networks (gan) to simulate
  central-place foraging trajectories,''
\newblock {\em Methods in Ecology and Evolution}, 2022.

\bibitem{traj_predict}
Y.~Ma, X.~Zhu, S.~Zhang, R.~Yang, W.~Wang, and D.~Manocha,
\newblock ``Trafficpredict: Trajectory prediction for heterogeneous
  traffic-agents,''
\newblock vol. 33, 2019.

\bibitem{gan_lagrangian}
J.~Gan, P.~Liu, and R.~K Chakrabarty,
\newblock ``Deep learning enabled lagrangian particle trajectory simulation,''
\newblock {\em Journal of Aerosol Science}, vol. 139, 2020.

\bibitem{cond_gan_traj}
S.~Julka, V.~Sowrirajan, J.~Schloetterer, and M.~Granitzer,
\newblock ``Conditional generative adversarial networks for speed control in
  trajectory simulation,''
\newblock in {\em International Conference on Machine Learning, Optimization,
  and Data Science}. Springer, 2021.

\bibitem{cond_gen_traj}
D.~Paz, H.~Zhang, and H.~I Christensen,
\newblock ``Tridentnet: A conditional generative model for dynamic trajectory
  generation,''
\newblock in {\em International Conference on Intelligent Autonomous Systems}.
  Springer, 2022.

\bibitem{traj_pred_rnn}
P.~Zegers and M.~K Sundareshan,
\newblock ``Trajectory generation and modulation using dynamic neural
  networks,''
\newblock {\em IEEE Transactions on Neural Networks}, vol. 14, 2003.

\bibitem{classif_dl_images}
Y.~Li, H.~Zhang, X.~Xue, Y.~Jiang, and Q.~Shen,
\newblock ``Deep learning for remote sensing image classification: A survey,''
\newblock {\em Wiley Interdisciplinary Reviews: Data Mining and Knowledge
  Discovery}, vol. 8, 2018.

\bibitem{cond_eulerian_lagr}
J.~F Quinting and C;~M Grams,
\newblock ``Eulerian identification of ascending airstreams (elias 2.0) in
  numerical weather prediction and climate models--part 1: Development of deep
  learning model,''
\newblock {\em Geoscientific Model Development}, vol. 15, 2022.

\bibitem{duong_geotracknet}
D.~Nguyen, R.~Vadaine, G.~Hajduch, R.~Garello, and R.~Fablet,
\newblock ``Geotracknet--a maritime anomaly detector using probabilistic neural
  network representation of ais tracks and a contrario detection,''
\newblock {\em IEEE Transactions on Intelligent Transportation Systems}, 2021.

\bibitem{lagrangian_flow}
R.~E Davis,
\newblock ``Lagrangian ocean studies,''
\newblock {\em Annual Review of Fluid Mechanics}, vol. 23, 1991.

\bibitem{seeding_radius_lagr_drift_error}
U.~Callies, M.~Kreus, W.~Petersen, and Y.~G Voynova,
\newblock ``On using lagrangian drift simulations to aid interpretation of in
  situ monitoring data,''
\newblock {\em Frontiers in Marine Science}, 2021.

\bibitem{traj_drifters_rnn}
N.~O Aksamit, T.~Sapsis, and G.~Haller,
\newblock ``Machine-learning mesoscale and submesoscale surface dynamics from
  lagrangian ocean drifter trajectories,''
\newblock {\em Journal of Physical Oceanography}, vol. 50, 2020.

\bibitem{markov_hidden}
W.~Zucchini and I.~L MacDonald,
\newblock {\em Hidden Markov models for time series: an introduction using R},
\newblock Chapman and Hall/CRC, 2009.

\bibitem{lagrangian_fçokker_planck_plankton}
A.~W Visser,
\newblock ``Lagrangian modelling of plankton motion: From deceptively simple
  random walks to fokker--planck and back again,''
\newblock {\em Journal of Marine Systems}, vol. 70, 2008.

\bibitem{PDE_Net}
Z.~Long, Y.~Lu, X.~Ma, and B.~Dong,
\newblock ``Pde-net: Learning pdes from data,''
\newblock in {\em International Conference on Machine Learning}. PMLR, 2018.

\bibitem{liu2011evaluation}
Y.~Liu and R.~H Weisberg,
\newblock ``Evaluation of trajectory modeling in different dynamic regions
  using normalized cumulative lagrangian separation,''
\newblock {\em Journal of Geophysical Research: Oceans}, vol. 116, 2011.

\bibitem{california_current}
R.~J Lynn and J.~J Simpson,
\newblock ``The california current system: The seasonal variability of its
  physical characteristics,''
\newblock {\em Journal of Geophysical Research: Oceans}, vol. 92, 1987.

\bibitem{elipot_drifters}
Lumpkin R. et~al. Elipot, S.,
\newblock ``A global surface drifter data set at hourly resolution,''
\newblock {\em Journal of Geophysical Research: Oceans}, vol. 121, 2016.

\bibitem{lellouche_glorys12}
J.M. Lellouche and E.~et~al. Greiner,
\newblock ``Recent updates to the copernicus marine service global ocean
  monitoring and forecasting real-time 1/ 12 high-resolution system,''
\newblock {\em Ocean Science}, vol. 14, 2018.

\bibitem{drifters_lumpkin}
R.~et~al. Lumpkin,
\newblock ``Evaluating where and why drifters die,''
\newblock {\em Journal of Atmospheric and Oceanic Technology}, vol. 29, no. 2,
  pp. 300--308, 2012.

\bibitem{lagr_time_scale}
W.~Krau{\ss} and C.~W B{\"o}ning,
\newblock ``Lagrangian properties of eddy fields in the northern north atlantic
  as deduced from satellite-tracked buoys,''
\newblock {\em Journal of Marine Research}, vol. 45, no. 2, pp. 259--291, 1987.

\bibitem{cond_gan_mirza}
M.~Mirza and S.~Osindero,
\newblock ``Conditional generative adversarial nets,''
\newblock {\em arXiv preprint arXiv:1411.1784}, 2014.

\bibitem{animal_traj_deep_learning}
T.~Maekawa, K.~Ohara, Y.~Zhang, M.~Fukutomi, S.~Matsumoto, K.~Matsumura,
  M.~Shidara, S.~J Yamazaki, R.~Fujisawa, K.~Ide, et~al.,
\newblock ``Deep learning-assisted comparative analysis of animal trajectories
  with deephl,''
\newblock {\em Nature communications}, vol. 11, no. 1, 2020.

\end{thebibliography}

\end{document}